\documentclass[a4paper,12pt,journal]{IEEEtran}
\usepackage[utf8]{inputenc}
\usepackage[T1]{fontenc}
\usepackage{fixltx2e}
\usepackage{graphicx}
\usepackage{grffile}
\usepackage{longtable}
\usepackage{wrapfig}
\usepackage{rotating}
\usepackage[normalem]{ulem}
\usepackage{amsmath}
\usepackage{textcomp}
\usepackage{amssymb}
\usepackage{capt-of}
\usepackage{url}
\usepackage{subfigure}
\usepackage{authblk}
\usepackage{float}
\usepackage{color}
\usepackage{hyperref}
\hypersetup{colorlinks,breaklinks,linkcolor=blue,urlcolor=blue,anchorcolor=blue,citecolor=blue}
\author[]{Eduardo Castelló Ferrer}
\affil[]{MIT Media Lab, 75 Amherst St., Cambridge, MA. \\ ecstll@mit.edu}
\date{}
\title{The blockchain: a new framework for robotic swarm systems}
\hypersetup{
 pdfauthor={eddie},
 pdftitle={The blockchain: a new framework for robotic swarm systems},
 pdfkeywords={},
 pdfsubject={},
 pdfcreator={Emacs 24.5.1 (Org mode 8.3.4)}, 
 pdflang={English}}
\begin{document}

\maketitle
\begin{abstract}

Swarms of robots will revolutionize many industrial applications, from
targeted material delivery to precision farming. However, several of
the heterogeneous characteristics that make them ideal for certain
future applications --- robot autonomy, decentralized control,
collective emergent behavior, etc. --- hinder the evolution of the
technology from academic institutions to real-world problems.

Blockchain, an emerging technology originated in the Bitcoin field,
demonstrates that by combining peer-to-peer networks with
cryptographic algorithms a group of agents can reach an agreement on a
particular state of affairs and record that agreement without the need
for a controlling authority. The combination of blockchain with other
distributed systems, such as robotic swarm systems, can provide the
necessary capabilities to make robotic swarm operations more secure,
autonomous, flexible and even profitable.

This work explains how blockchain technology can provide innovative
solutions to four emergent issues in the swarm robotics research
field. New security, decision making, behavior differentiation and
business models for swarm robotic systems are described by providing
case scenarios and examples. Finally, limitations and possible future
problems that arise from the combination of these two technologies are
described.

\end{abstract}

\section{The blockchain: a disruptive technology}
\label{sec:orgheadline1}

\begin{figure}[thbp] 
    \begin{center}
      \subfigure[]
                {\includegraphics[width=0.45\textwidth]{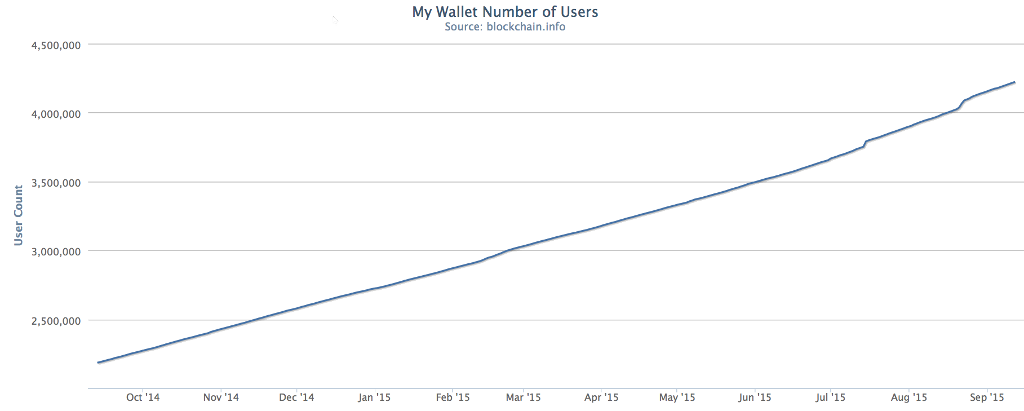}
                 \label{fig:BitCoinUsers}
                } 
      \subfigure[]
                {\includegraphics[width=0.45\textwidth]{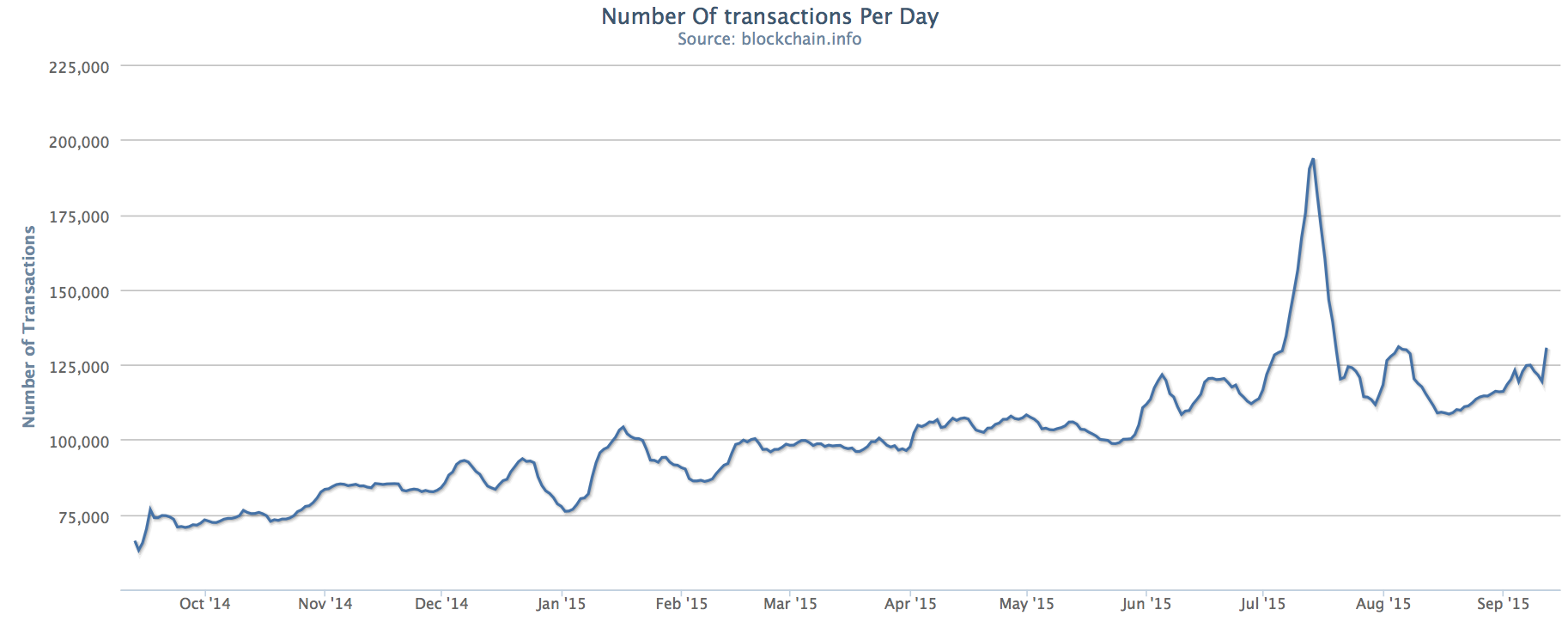}
                  \label{fig:BitCoinTransactions}
                }
      \label{fig:BitCoinStats}
      \caption{ (a) Total number of users of the most popular
      Bitcoin client --- MyWallet --- during the Sep 2014 - Sep 2015 period.
      (b) Total number of Bitcoin transactions during the Sep 2014 -
      Sep 2015 period.}
    \end{center}
  \end{figure}

In September 2008, Satoshi Nakamoto introduced two influential ideas
in his white paper ``Bitcoin: A Peer-to-Peer Electronic Cash
System''\footnote{\url{http://bitcoin.org/bitcoin.pdf}}. The first was ``Bitcoin'' --- a
decentralized, peer-to-peer, online currency able to maintain value
without any backing from a central authority. After garnering an
increasing amount of attention from early adopters
\cite{BitCoinEarlyAdopterWired2013} and law makers \cite{Brito2013},
Bitcoin became recognized as a cheap, rapid, and reliable method of
moving economic value across the internet in a decentralized
manner. With over 4 million users as seen in
Fig. \(\ref{fig:BitCoinUsers}\)\footnote{Source: \url{http://blockchain.info/charts/my-wallet-n-users}}, and over 125,000
transactions per day as seen in
Fig. \(\ref{fig:BitCoinTransactions}\)\footnote{Source: \url{http://blockchain.info/charts/n-transactions-total}}, Bitcoin
has transformed into one of the most powerful computing networks in
existence \cite{ForbesComputingPower2015}.

\begin{figure}[htb]
\centering
\includegraphics[width=0.5\textwidth]{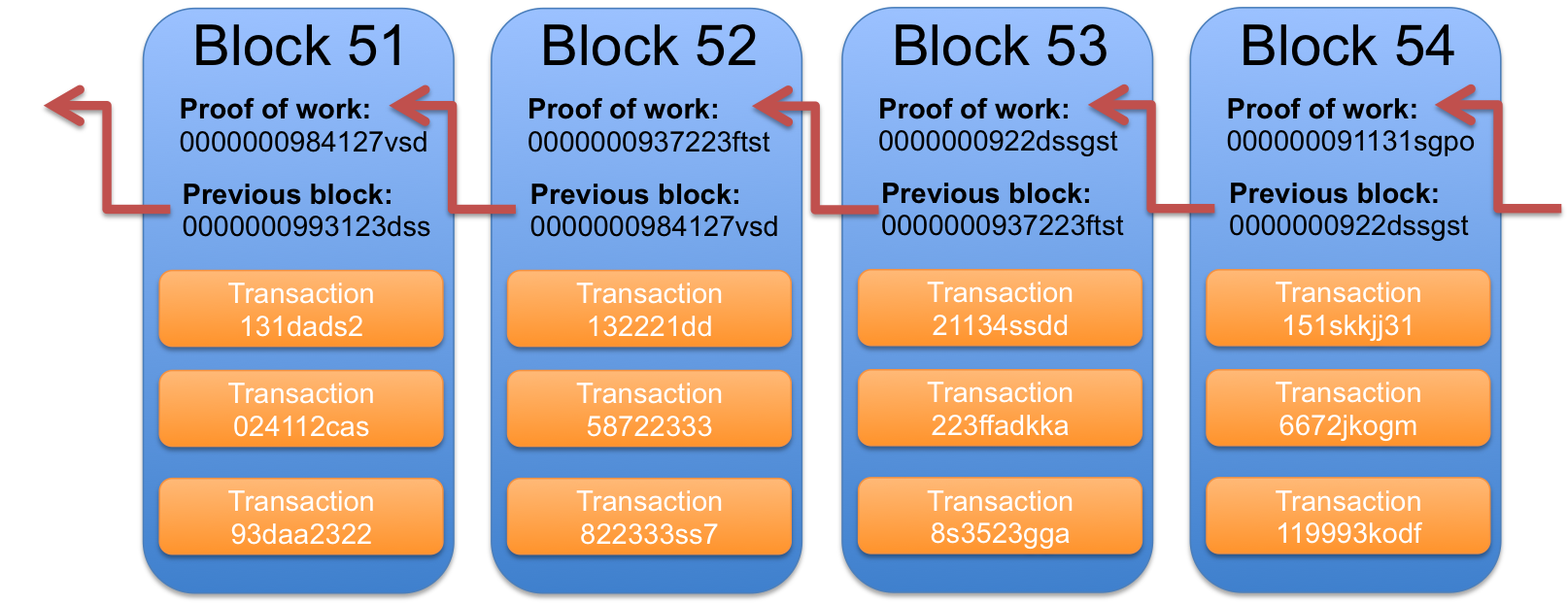}
\caption{\label{fig:orgparagraph1}
A simple section of a blockchain}
\end{figure}

The second, equally important idea was the ``blockchain'', which is a
public chronological database of transactions recorded by a network of
agents. Individual transactions containing details of who sent what to
whom are grouped into datasets referred to as ``blocks'', as illustrated
in Fig. \ref{fig:orgparagraph1}.

Each block contains information about a certain number of
transactions, a reference to the preceding block in the blockchain,
and an answer to a complex mathematical challenge known as the ``proof
of work''. The concept of proof of work is used to validate the data
associated with that particular block as well as to make the creation
of blocks computationally ``hard'', thereby preventing attackers from
altering the blockchain in their favor\footnote{Recently, new techniques, such as ``proof of
stake'', requiring no computational work to validate
blocks, have been introduced to expand blockchain technology to
resource-limited devices. More information about ``proof of stake''
systems can be found in: \url{http://peercoin.net/assets/paper/peercoin-paper.pdf}}. It is based
on cryptographic techniques --- SHA256 in the case of Bitcoin ---, which
output unpredictable numeric values, also known as hashes, that
encapsulate all transactions within a block in a digital
fingerprint. Any differences in the input data --- transaction order,
quantities, receivers, etc. --- will produce differences in the output
data --- proof of work hash --- and, thus, a different digital
fingerprint.

\begin{figure}[htb]
\centering
\includegraphics[width=0.5\textwidth]{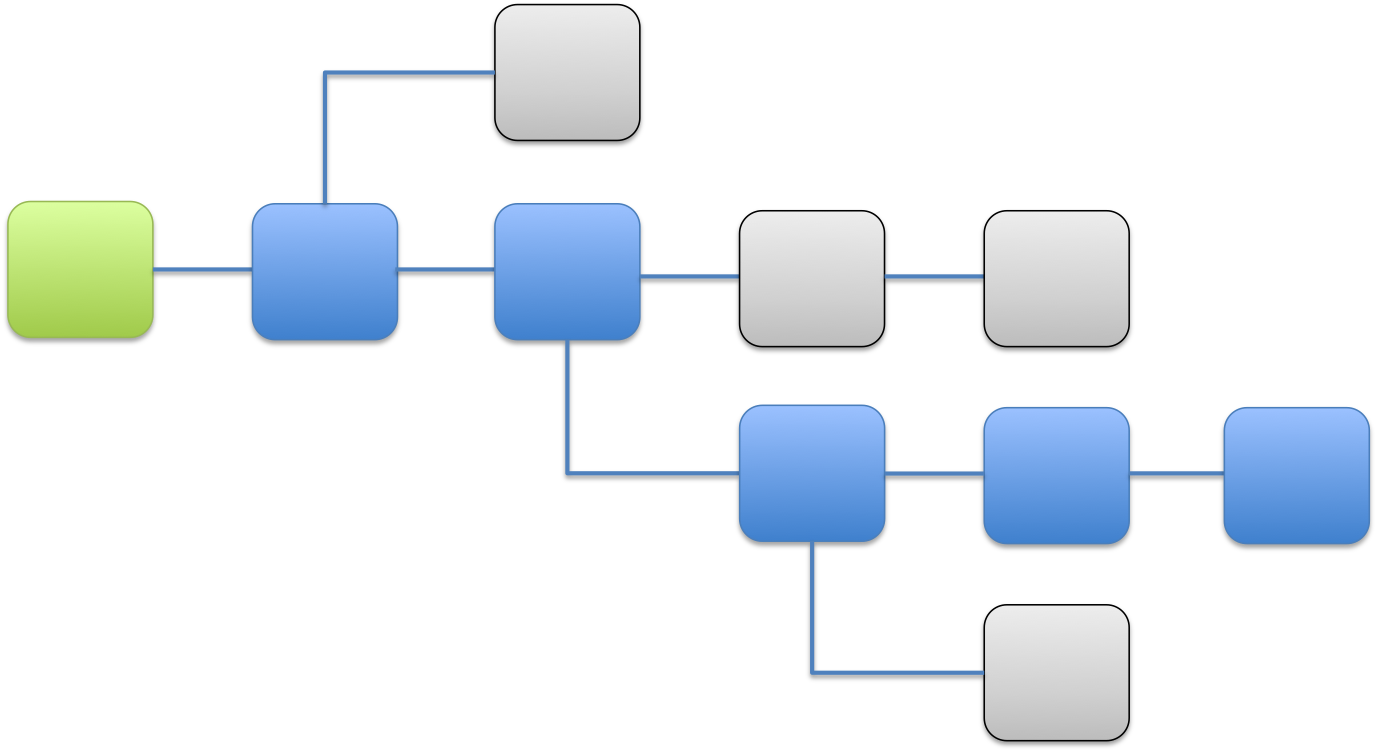}
\caption{\label{fig:orgparagraph2}
A graphical representation of the blockchain}
\end{figure}

After ensuring that all new transactions to be included in the block
are valid and do not invalidate previous transactions, e.g., through
double-spending, a new block is added to the end of the blockchain by
an agent in the network, hereafter referred to as a miner. At that
moment, the information contained in the block can no longer be
deleted or modified, and it is available to be certified by everyone
on the network. A copy of the blockchain, similar to the one
illustrated in Fig. \ref{fig:orgparagraph2}, is stored by
every agent and is periodically synchronized in a peer-to-peer fashion
to ensure that they all share the same public database. With these
properties, the blockchain becomes a permanent record that all agents
on the network can use to coordinate an action, verify an event, and
reach an agreement in an auditable way without the need for a
centralized authority. However, due to its decentralized nature, the
blockchain sometimes produces orphaned blocks, depicted by the grey
blocks in Fig. \ref{fig:orgparagraph2}, which occur naturally
when two miners produce a block at a similar time. Initially accepted
by a part of the network, these blocks are later rejected when proof
of a longer blockchain is received.

Several projects are currently exploring the potential benefits of
blockchain technology in a wide range of sectors such as intellectual
property, real estate, etc. \cite{Prisco2014}. Beyond this, two of
the most promising projects concerning blockchain technology are
Bitcongress\footnote{\url{http://bitcongress.org/}} and Colored
Coins\footnote{\url{http://coloredcoins.org/}}. Bitcongress is a decentralized voting platform
intended for nations, states, or communities, to ease the legislation
and rule-making process by providing a secure and auditable voting
system. The Colored Coins project is focused on creating digital
assets that can represent real-world value. By attaching metadata to
Bitcoin transactions, digital tokens on the blockchain can be used to
store information --- documents, certificates, etc. ---, provide prove
of ownership rights, or issue financial assets such as shares. Due to
the latter, ``colored coins'' can be used to create Distributed
Collaborative Organizations (DCO), which are basically virtual
entities with shareholders. In those situations, the blockchain helps
to keep track of a company's ownership structure, as well as to create
and distribute shares for DCOs in a transparent and secure way.

Blockchain technology demonstrates that by combining peer-to-peer
networks with cryptographic algorithms, a group of agents can reach an
agreement on a particular state of affairs and record that agreement
in a secure and verifiable manner without the need for a controlling
authority. Due to its decentralized nature, and key underlying
principles such as robustness and fault-tolerance, blockchain
technology may be useful in combination with emergent fields including
automated transportation, logistic and warehouse systems or even cloud
computing. The aim of this work is to outline the potential benefits
of combining blockchain technology with robotics --- specifically,
swarm robotics and state-of-the-art robotic hardware, which have
garnered increasing attention in both academic and industrial sectors
--- and to emphasize how this synergy can ease the transition from
academic research to real-world applications and eventual widespread
industrial use.

\section{Swarm Robotics: The emergent field}
\label{sec:orgheadline6}

\begin{figure}[htb]
\centering
\includegraphics[width=0.45\textwidth]{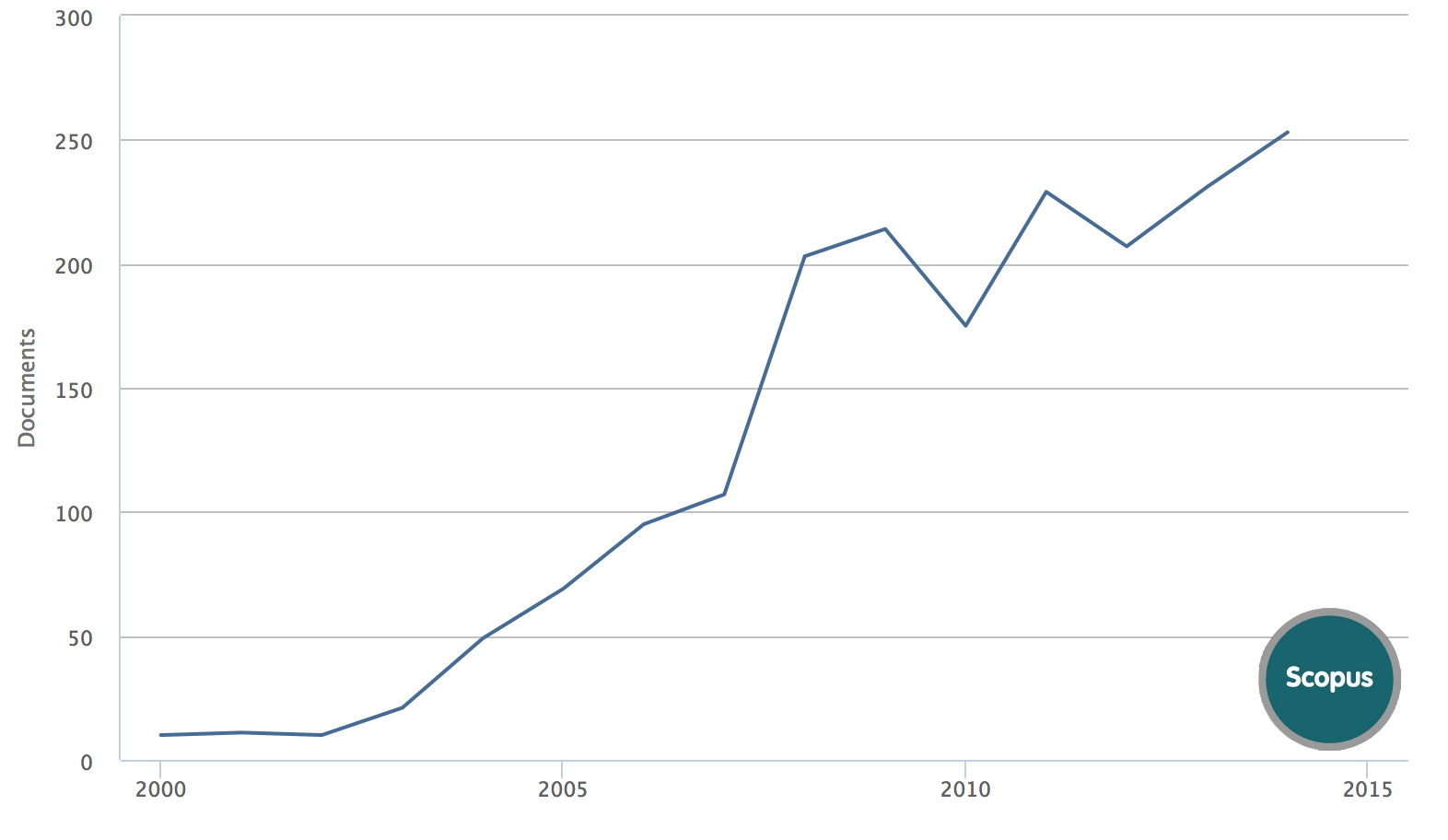}
\caption{\label{fig:orgparagraph3}
Total number of research documents on swarm robotic systems published annually from 2000 - 2014.}
\end{figure}

With a strong initial influence from nature and bio-inspired models
\cite{Momen2010,Walker2011,Bonabeau1999}, swarm systems are known for
their adaptability to different environments \cite{Bentes2012} and
tasks \cite{Brambilla2013}. Key advantages of robotic swarms are
robustness to failure and scalability, both of which are due to the
simple and distributed nature of their coordination. Due to these
characteristics, global behaviors are not explicitly stated, and
instead emerge from local interactions between robots. As a result,
swarm robotics research has recently been gaining popularity, as
demonstrated in Fig. \ref{fig:orgparagraph3}\footnote{Source: Scopus research database.}.

As the cost of robotic platforms continues to decrease, the number of
applications involving robot swarms is increasing. These include
targeted material transportation \cite{Chen2013}, where groups of
small robots are used to carry tall and potentially heavy objects,
precision farming \cite{Emmi2014,Yaghoubi2013}, where a fleet of
autonomous agents shift operator activities in agricultural tasks, and
even entertainment systems \cite{Alonso-Mora2014,Hortner2012}, where
multiple robots come together to form interactive displays. Several
breakthroughs originating in this field have had a direct impact on
the emergence of technologies such as Unmanned Aerial Vehicles (UAVs)
\cite{Cekmez2014,Varela2011} and nanorobotics
\cite{Kaewkamnerdpong2009,Nantapat2011,Chandrasekaran2006}.

\begin{figure}[thbp] 
\begin{center}
      \subfigure[]
      {\includegraphics[width=0.48\textwidth]{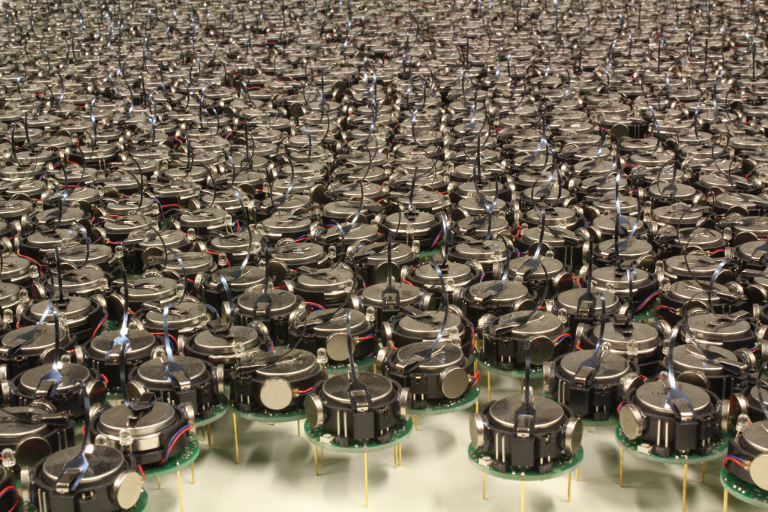}
      \label{fig:KilobotSwarm}
      } 
      \subfigure[]
      {\includegraphics[width=0.48\textwidth]{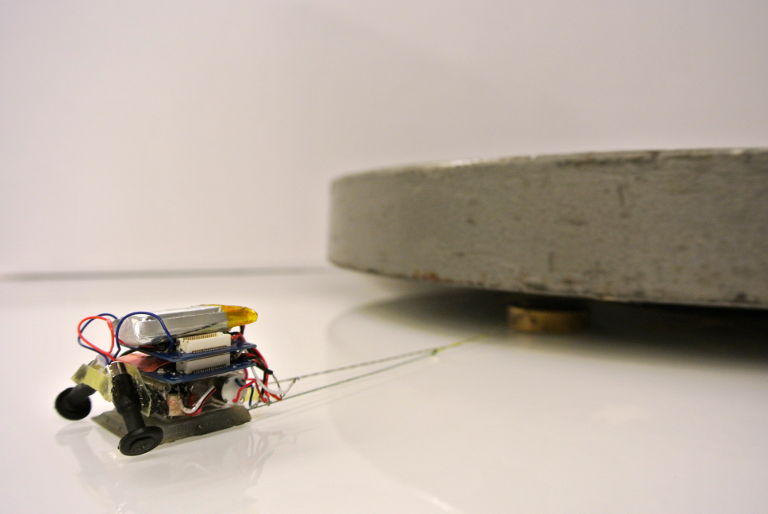}
      \label{fig:MicroTug}
      }
      \label{fig:EmergentSwarmPlatforms}
      \caption{ (a) A swarm of 1024 Kilobot \cite{Rubenstein2012}
      robots. The Kilobot robot demonstrated that a low-cost
      platform --- $\$$14 worth of parts --- can be a viable solution for
      producing swarms of hundreds or thousands of members. (b)
      Microtug robot carrying a weight. Microtug robot
      \cite{Christensen2015} towing a weight. Microtug robots use
      an innovative adhesive technology to move 2000 times their
      own weight.}
\end{center}
\end{figure}

These examples, along with the growing development of robotic hardware
\cite{Rubenstein2012,Christensen2015}, suggest that commercial
applications for swarms are within reach. However, as new swarm
robotics companies \cite{DAndrea2012,Bonirob2009} have started to
emerge, it is clear that there are problems in effectively
transferring knowledge and technology from academic institutions to
the industry \cite{Bayndr2015}. Previous works have emphasized the
lack of general methods to tackle topics such as safety analysis,
testing mechanisms \cite{Winfield2005,Csahin2008} or security
protocols \cite{Higgins2009} for swarm robotic systems, which hinder
the progress to more broader commercial applications.

One of the main axioms in the swarm robotics field has been the
absence of global knowledge or explicit communication models between
swarm robots. Traditionally, swarm robotic systems exclusively rely on
local communication --- e.g., between adjacent robots in a flocking
mission ---, and no global knowledge is maintained within the swarm.
Therefore, the use of blockchain technology in combination with swarm
robotic systems might be seen as a diversion from the main research
approach. However, the use of global knowledge in swarm robotic
systems has been proved useful for different applications such as
cooperative techniques to cope with unknown environments
\cite{Jamshidpey2015} or the synchronization between different swarm
teams \cite{Majercik2012}.

These findings suggest that the combination of both types of
information --- local and global --- might co-exist \cite{Bayndr2015}
without compromising the robustness to failure and scalability
properties of these systems. In addition, recent achievements in
hardware design and manufacturing, such as the Raspberry
Pi\footnote{\url{http://www.raspberrypi.org/}} or Intel Galileo\footnote{\url{http://www.intel.com/content/www/us/en/embedded/products/galileo/galileo-overview.html}} motherboards,
enable nowadays robots to count with increasing processing
capabilities as well as low-power communication devices. These
advancements open the door to include explicit communication and
global knowledge models in swarm robotic systems.

In the following, I will discuss how blockchain and its underlying
principles can be useful for tackling four emergent issues in the
swarm robotics field by using the robots as nodes in a network and
encapsulating their transactions in blocks.

\subsection{Security}
\label{sec:orgheadline2}

One of the main obstacles to the large-scale deployment of robots
for commercial applications is security. Previous research has
highlighted the necessity of developing systems in which swarm
members can detect and trust their counterparts
\cite{Zikratov2014}. This is especially important, since it has
been demonstrated that the inclusion of swarm members that are
``faulty'' or have malicious intentions could be a potential risk for
the swarm's goals \cite{Millard2014} as well as a security breach.

Security in any environment, including swarm robotic systems, is
fundamentally about the provision of core services such as data
confidentiality, data integrity, entity authentication, and data
origin authentication. In contrast to other fields in which
security-related research is being actively conducted, swarm
robotic systems suffer a lack of practical solutions to these
problems \cite{Higgins2009}. The security topic has been
overlooked by state-of-the-art research mainly due to the complex
and heterogeneous characteristics of robotic swarm systems ---
robot autonomy, decentralized control, a large number of members,
collective emergent behavior, etc. Technology such as blockchain
can provide not only a reliable peer-to-peer communication channel
to swarm's agents, but are also a way to overcome potential
threats, vulnerabilities, and attacks.

\begin{figure}[htb]
\centering
\includegraphics[width=0.5\textwidth]{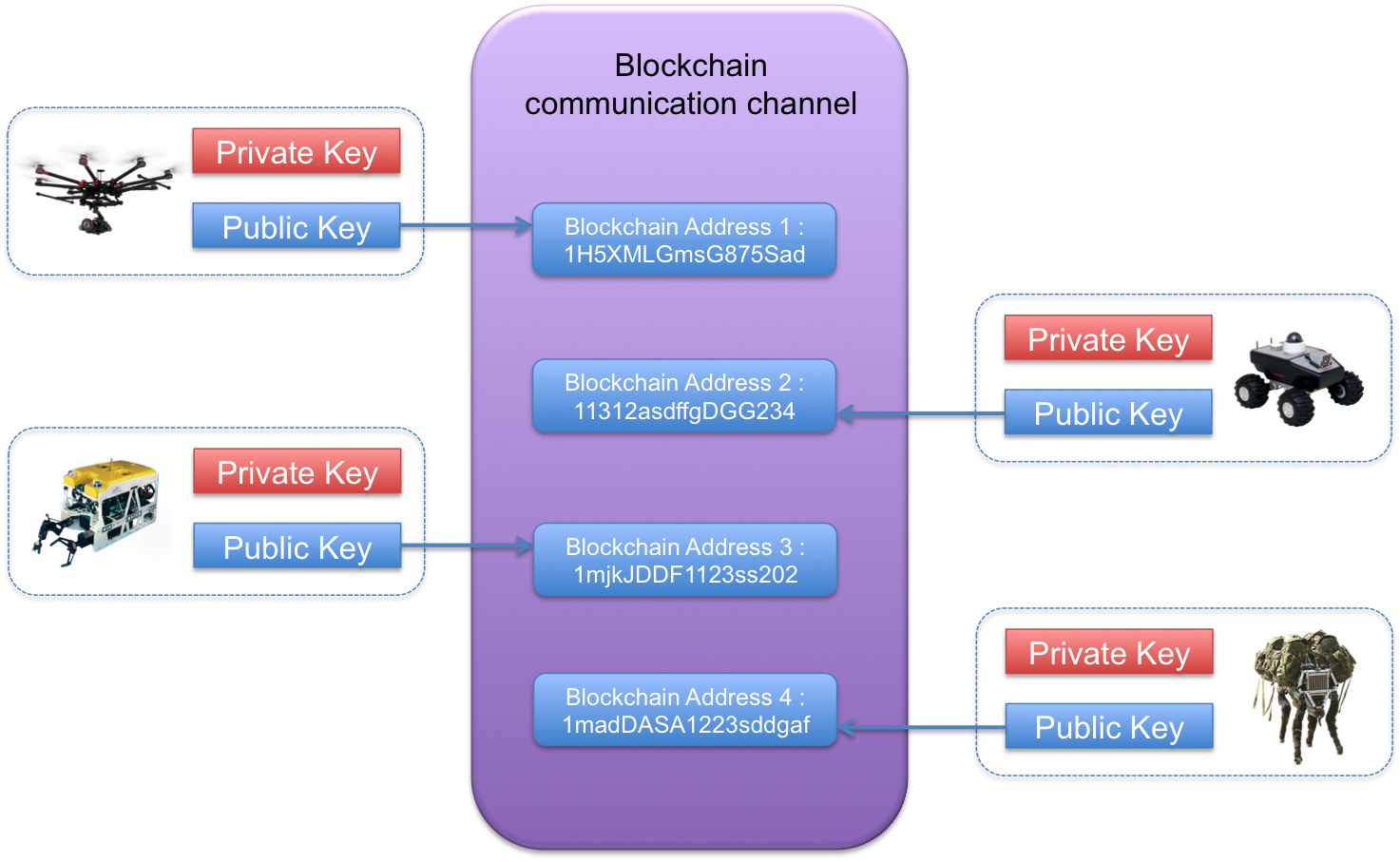}
\caption{\label{fig:orgparagraph4}
Different types of robots share the blockchain communication channel using their public keys as main identifiers.}
\end{figure}

In the blockchain encryption scheme, techniques such as public key
and digital signature cryptography are accepted means of not only
making transactions using unsafe and shared channels, but also of
proving the identity of specific agents in a network. A pair of
complementary keys, public and private, are created for each agent
to provide these capabilities, as illustrated in
Fig. \ref{fig:orgparagraph4}. Public keys are an agent's
main accessible information, are publicly available in the
blockchain network, and can be regarded as a special type of
account number. In contrast, private keys are an agent's secret
information --- similar to passwords in traditional systems --- and are
exclusively used to validate an agent's identity and the operations
that it may execute.

\begin{figure}[htb]
\centering
\includegraphics[width=0.5\textwidth]{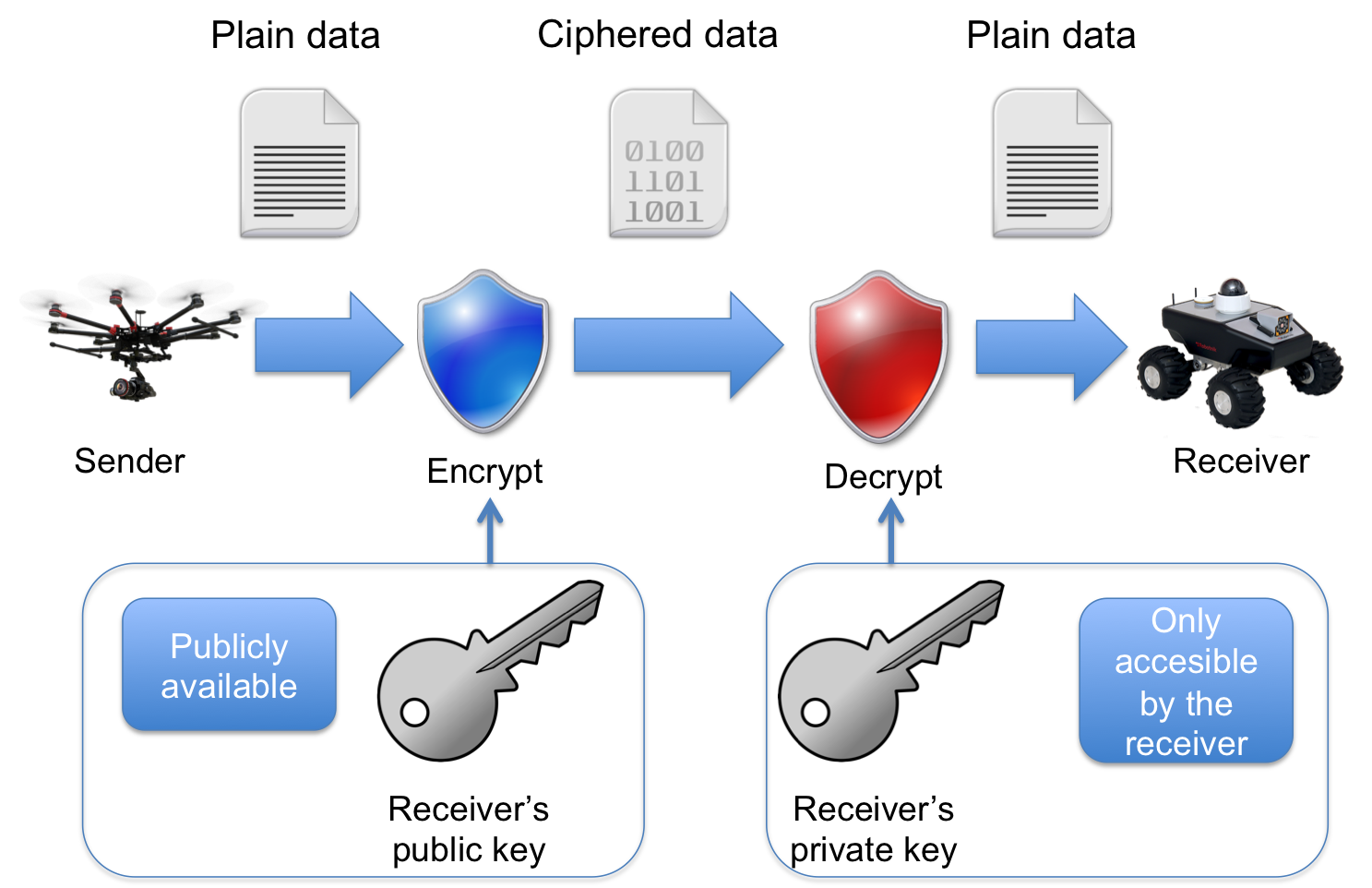}
\caption{\label{fig:orgparagraph5}
Public key cryptography allows robots to be sure that the content of a message can only be read by the owner of the corresponding sending address.}
\end{figure}

In the case of swarm robotics, public key cryptography as depicted
in Fig. \ref{fig:orgparagraph5} allows robots to share their
public keys with other robots who want to communicate with
them. Therefore, any robot in the network can send information to
specific robot addresses, knowing that only the robot that
possesses the matching private key can read the message. Since the
public key cannot be used to decrypt messages, there is no risk if
it falls into the wrong hands. In addition, it prevents third-party
robots from decrypting such information even if they share the same
communication channel.

\begin{figure}[htb]
\centering
\includegraphics[width=0.5\textwidth]{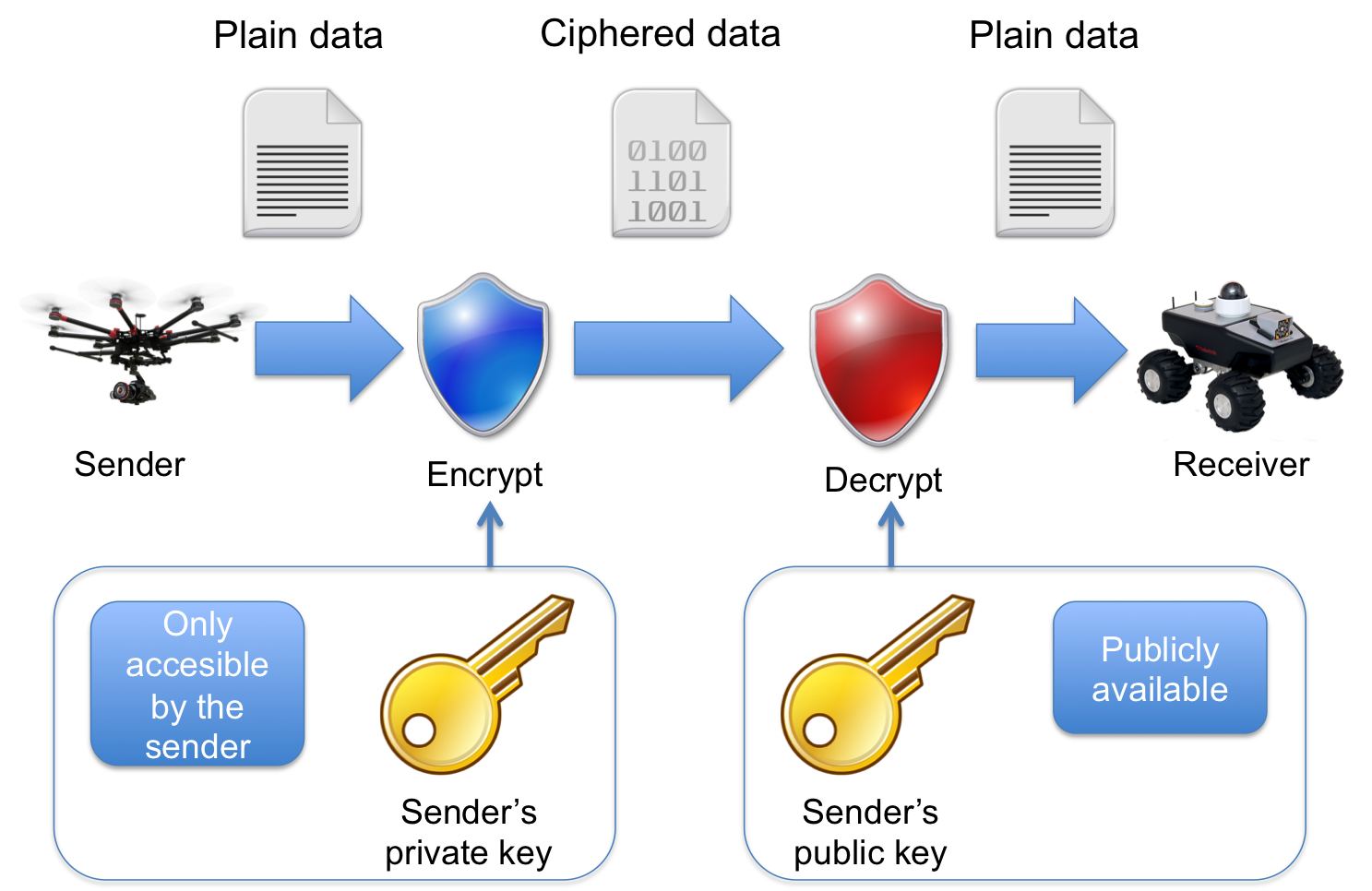}
\caption{\label{fig:orgparagraph6}
Digital signature cryptography provides a way to prove the ownership of a specific address --- public key.}
\end{figure}

Complementing the above, digital signature cryptography, as
illustrated in Fig. \ref{fig:orgparagraph6}, allows robots
to use their own private key to encrypt messages. Other robots can
then decrypt them using the sender's public key. As any robot has
access to the sender's public key, the contents of the message will
not be a secret, but the fact that it was encrypted using the
sender's private key proves that the message could not have been
sent by anyone else, thereby proving its authorship.

On one hand, public key cryptography ensures that the content of a
message --- encapsulated in a blockchain transaction, for instance ---
can only be read by the robot owning a specific address. On the
other hand, digital signature cryptography can provide entity
authentication and data origin authentication between robots or
third-party agents.

Applications that can potentially benefit from the security
features provided by blockchain technology include the military
\cite{McCook2007}, where the need for reliable and trustworthy
systems is self-evident, and disaster relief
\cite{Liu2013,Stormont2005}, where accurate identification between
aid agencies is crucial, especially in the case where multiple
swarms are in joint operation \cite{Stormont2003}. Another relevant
example is the I-ward project \cite{Thiel2009}, in which robot teams
provide assistance to health-care workers in the transportation of
medicines and patients' medical records. In this case, entity
authentication and data confidentiality may be the most important
security requirements.

\subsection{Distributed decision making}
\label{sec:orgheadline3}

Distributed decision making algorithms have played a crucial role
in the development of swarm systems. One of the most prominent
examples is the use of robot swarms connected through ad-hoc
networks --- MANET --- \cite{Derr2013,Yong2009} to achieve
distributed sensing applications. These systems have the capability
to sense information from multiple viewpoints and, thus, increase
the quality of data obtained. However, the robots in the swarm need
to reach a global agreement regarding the object of interest ---
e.g., paths to traverse, shape to form, or obstacles to avoid.
Hence, there is a need to develop distributed decision making protocols
\cite{Li2011} that ensure guaranteed convergence towards a common
outcome.

Distributed decision making algorithms have been adopted in many
robotic applications, including dynamic task allocation
\cite{Das2011}, collective map building \cite{Aragues2012}, and
obstacle avoidance \cite{Navarro2011}. However, the deployment of
large quantities of agents with distributed decision-making is
still an open problem \cite{Pourmehr2013}. Several well-known
trade-offs, such as speed versus accuracy during collective
decision-making processes, have been identified
\cite{Franks2003,Wei2005,Valentini2015}, and are a key aspect for
consideration before real-world deployments. Therefore, more
autonomous and flexible solutions to robot decision making in
distributed systems are required to tackle the new wave of
challenges facing the industry. Blockchain is an outstanding
technology for ensuring that all participants in a decentralized
network share an identical view of the world. For instance,
blockchains allow for the possibility of creating distributed
voting systems for robot swarms that need to reach an agreement.

\begin{figure}[thbp] 
\begin{center}
      \subfigure[]
      {\includegraphics[width=0.48\textwidth]{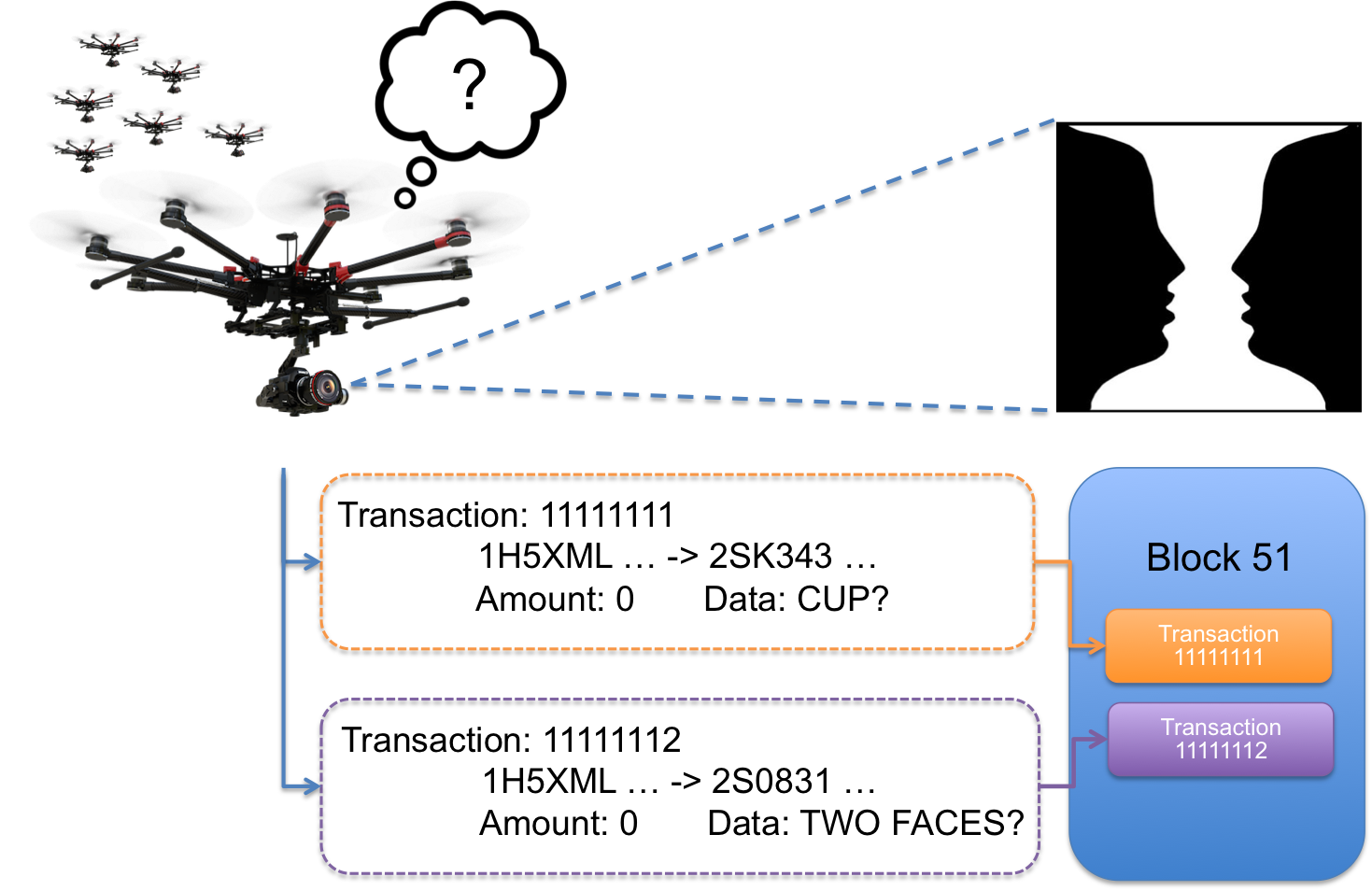}
      \label{fig:VoteCreation}
      }
      \subfigure[]
      {\includegraphics[width=0.48\textwidth]{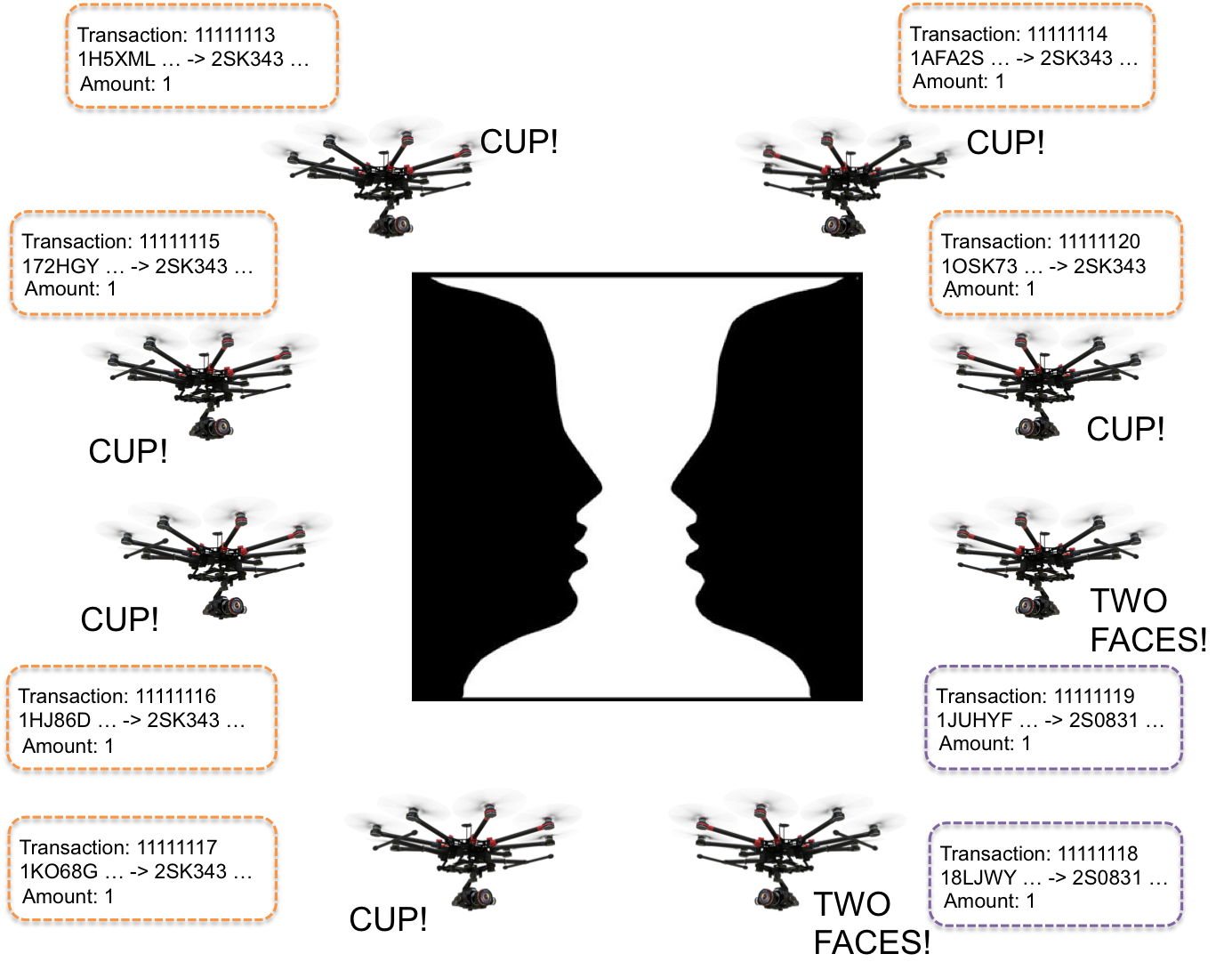}
      \label{fig:VoteProcess}
      }
      \subfigure[]
      {\includegraphics[width=0.48\textwidth]{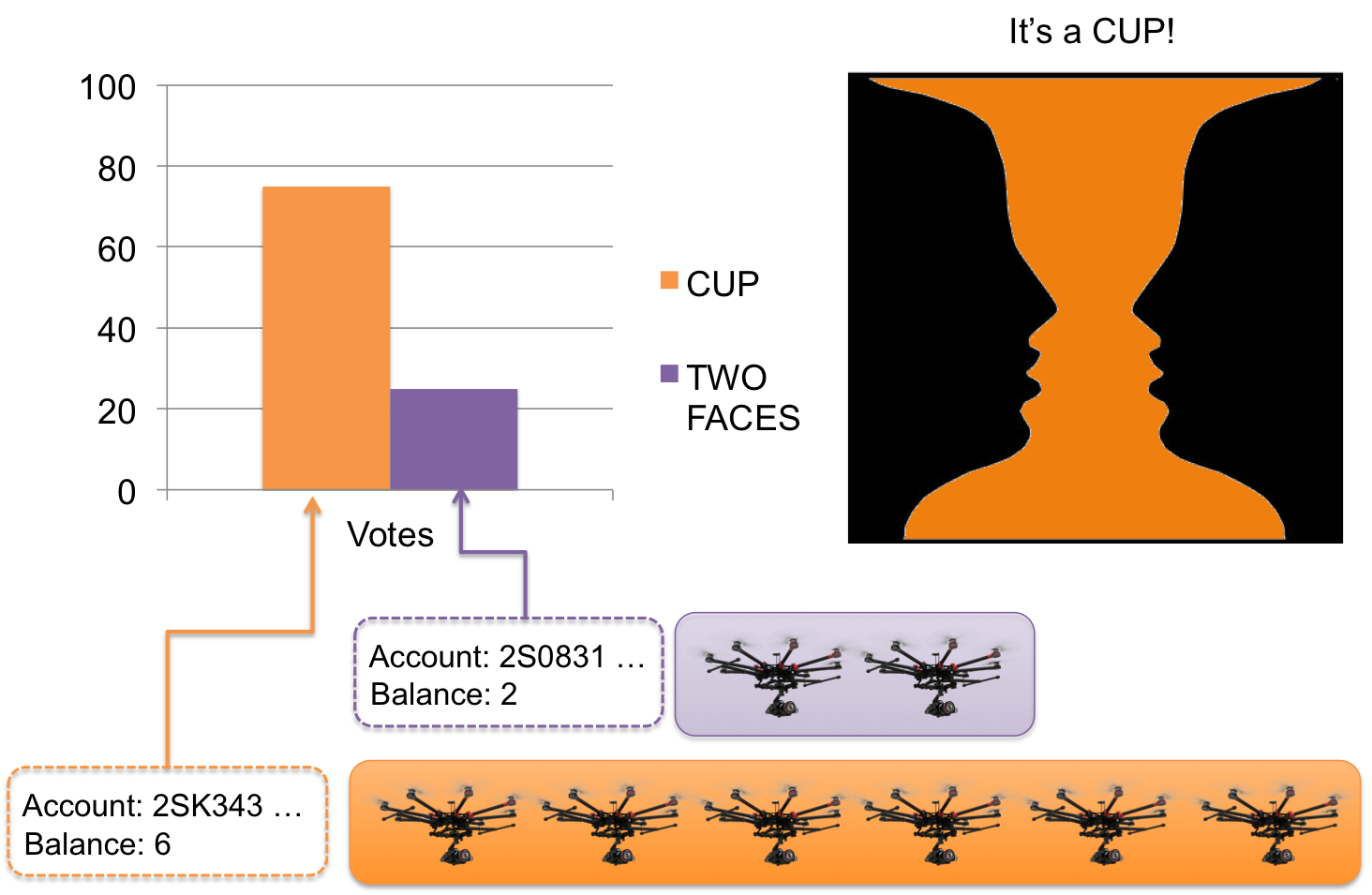}
      \label{fig:VoteResolution}
      }
      \caption{(a) One of the swarm members recognizes an object of
      interest during the mission. In order to reach an agreement
      the swarm robot executes two transactions, creating two
      special addresses representing the possible options and
      registering them in the blockchain. (b) The rest of the swarm
      gathers around the object to obtain different
      perspectives. Each swarm member issues a new transaction to
      the account matching their classification algorithm. (c) When
      the voting process ends, the entire swarm reaches an
      agreement about the object based on the voting results.}
      \label{fig:BlockChainVote}
\end{center}
\end{figure}

Figure \ref{fig:BlockChainVote} outlines a simple example of how
blockchain technology can be used to assist in the decision making
process of robotic swarms. Every time a swarm member is in a
situation requiring an agreement, it can issue a special
transaction, creating an address associated with each of the
possible options the robotic swarm has to choose from, as shown in
Fig. \ref{fig:VoteCreation}. After being included in a block, the
information is publicly available and other swarm members can vote
according to their situation by, for example, transferring one
token to the address corresponding to their chosen option, as shown
in Fig. \ref{fig:VoteProcess}. Agreements --- e.g., by the majority
rule --- can be obtained rapidly and in a secure and auditable way
since all robots can monitor the balance of addresses involved in
the voting process as shown in Fig. \ref{fig:VoteResolution}.

Furthermore, the inclusion of blockchain technology in robotic
swarms opens the path to achieving more advanced collaborative
models between robots using multi-signature (multisig) techniques.
Multisig techniques rely on addresses and transactions that are
associated with more than one private key. The simplest type of
multisig address is called an m-of-n address --- where m \(<\) n ---
, which is an address associated with n private keys that requires
signatures from at least m keys to transfer information. Complex
collaborative missions especially designed for heterogeneous groups
of robots are easy to formalize, publish, and carry out in this
way.

\begin{figure}[thbp] 
\begin{center}
      \subfigure[]
      {\includegraphics[width=0.45\textwidth]{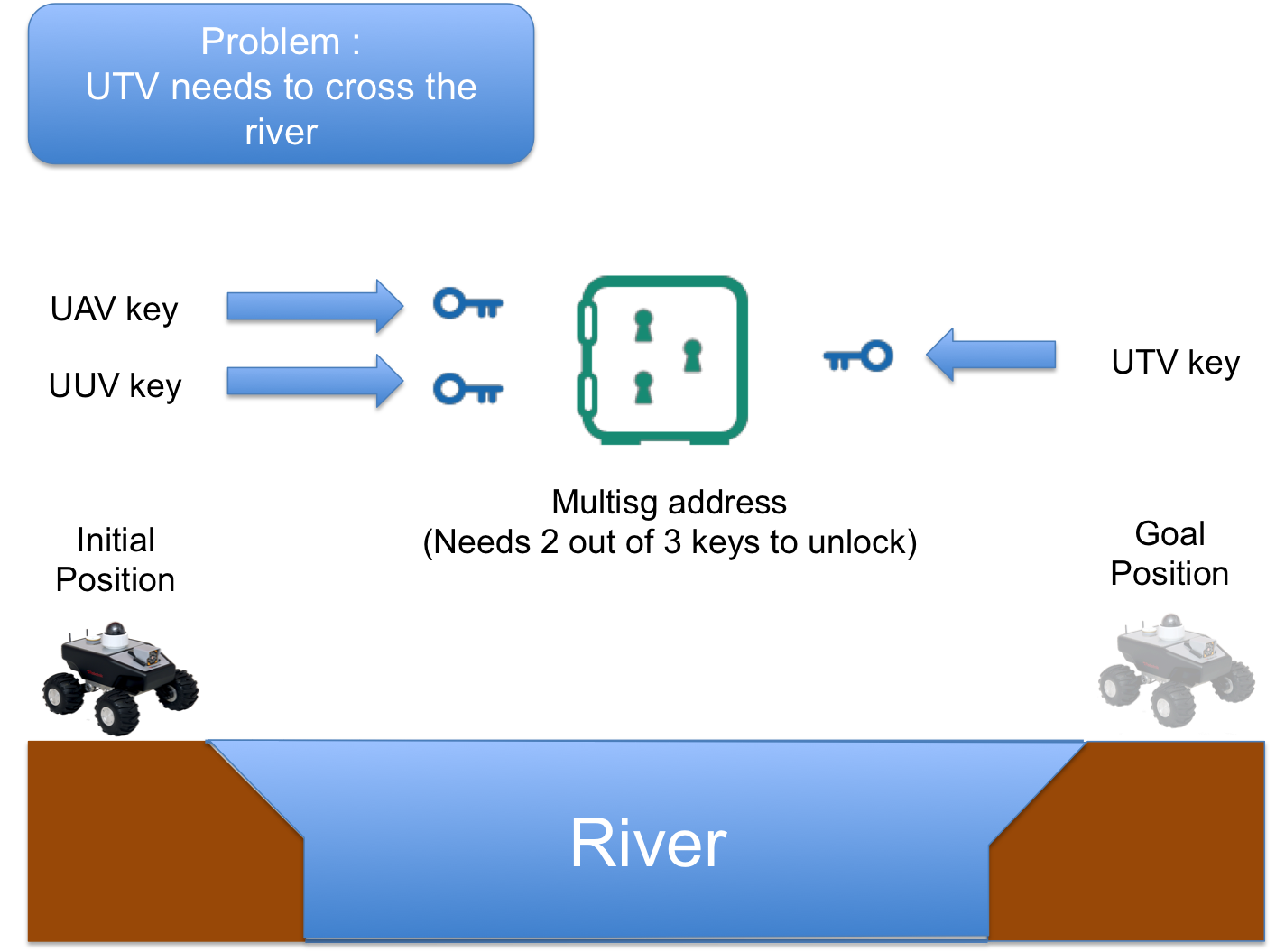}
      \label{fig:MultiSignProblem}
      } 
      \subfigure[]
      {\includegraphics[width=0.45\textwidth]{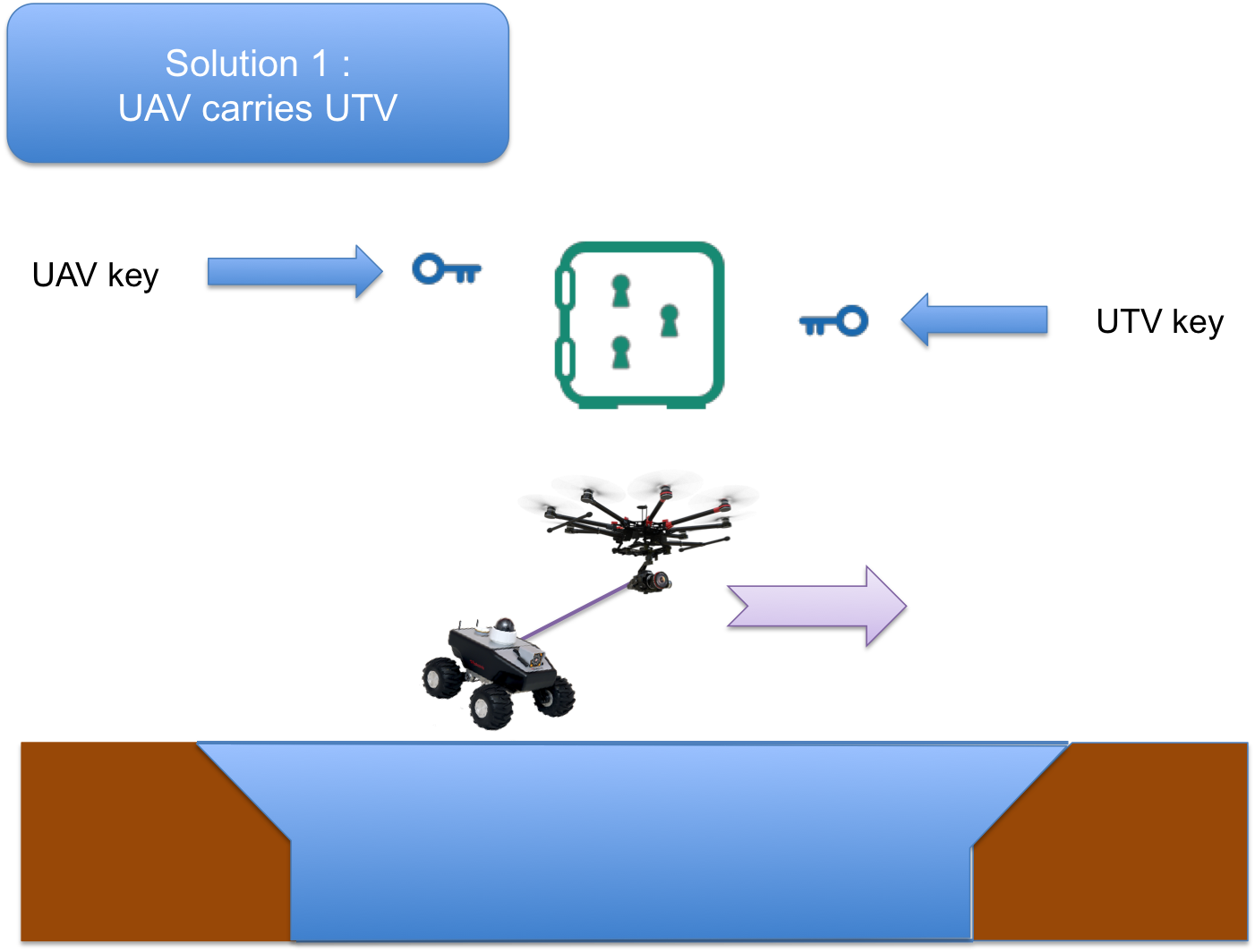}
      \label{fig:MultiSignSolution1}
      }
      \subfigure[]
      {\includegraphics[width=0.45\textwidth]{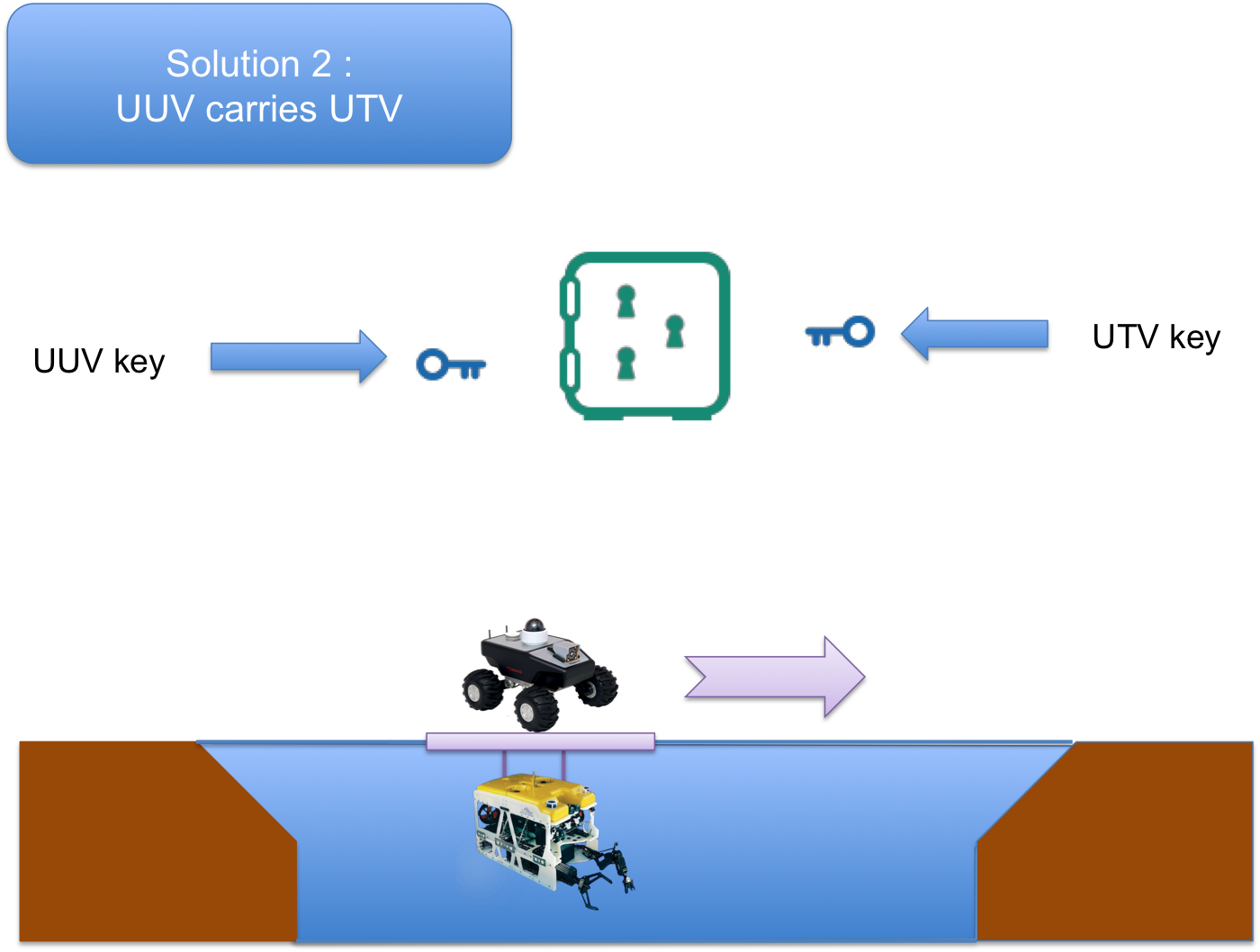}
      \label{fig:MultiSignSolution2}
      }
      \caption{An Unmanned Terrestrial Vehicle (UTV) faces the
      problem of crossing a river bed. It creates a multisig
      address in which 2-of-3 keys are needed to establish the
      collaboration and solve the problem. (b) Possible Solution 1:
      An available Unmanned Aerial Vehicle (UAV) provides its key
      to unlock the multisig address and retrieve the UTV
      position. The problem is solved by permitting the UAV to
      carry the UTV to the other side. (c) Possible Solution 2: An
      available Unmanned Underwater Vehicle (UUV) provides its key
      to unlock the multisig address and retrieve the UTV's
      position. The action is fulfilled by letting the UUV carry
      the UTV to the other side.}
      \label{fig:BlockChainMultiSign}
\end{center}
\end{figure}

Figure \ref{fig:BlockChainMultiSign} provides a simple overview of
the potential capabilities of multisig addresses in swarm
collaborative missions. In this case, an Unmanned Terrestrial
Vehicle (UTV) with the need to avoid an obstacle --- a river ---
can create a partially signed transaction representing a call for
assistance, as shown in Fig. \ref{fig:MultiSignProblem}, and
distribute it across the network. At that moment, a suitable robot
unit such as an Unmanned Aerial Vehicle (UAV), as shown in
Fig. \ref{fig:MultiSignSolution1}, or an Unmanned Underwater
Vehicle (UUV), as shown in Fig. \ref{fig:MultiSignSolution2}, can
sign its part of the transaction responding to the call. This
action will unlock information such as the UTV's position and even
the tokens contained within the multisig address as payment to
complete the action. Under this collaboration scheme, more
autonomous and emergent behaviors can arise within the robot
swarm. For instance, robots in unfavorable circumstances --- e.g., low
battery levels, poor sensor readings, etc. --- could be more reactive
to requests for assistance from other robots who provide valid
tokens, doing so to improve their own situation within the
swarm. Robots could purchase battery refills, obtain higher-quality
sensor readings, or simply request other services from other robots
in order to maximize their own, personal goals.

Finally, the adoption of blockchain technology in the distributed
decision processes of robotic swarms can provide additional benefits to the
robotic swarms' maintainers and operators. Due to the fact that all
agreements and all related transactions are stored in the
blockchain, there is no need to invest time in learning and
training phases for new robots joining the swarm. Instead, these
new robots will be able to automatically synchronize with the rest
of the swarm by downloading the ledger containing the history of
all agreements and knowledge previously discovered and stored in
the blockchain.

\subsection{Behavior differentiation}
\label{sec:orgheadline4}

The combination of blockchain technology with classical swarm
control techniques can be useful in tackling problems beyond
security and distributed decision making issues. According to
recent surveys \cite{Bayndr2015,Brambilla2013}, even though
state-of-the-art algorithms have enabled specialized teams of
robots to handle individual specific behaviors --- aggregation,
flocking, foraging, etc. ---, robot swarms deployed in the real
world will likely need to handle a number of different behaviors,
for example, by switching from one control algorithm to another to
accomplish a given objective. The combination of different
behaviors in a swarm has not been diligently studied in the
literature \cite{Bayndr2015}. However, blockchain technology
provides the possibility of linking several blockchains in a
hierarchical manner, also known as pegged sidechains\footnote{\url{http://www.blockstream.com/sidechains.pdf}},
which would allow robotic swarm agents to act differently according
to the particular blockchain being used, where different
parameters, such as mining diversity, permissions, etc., can be
customized for different swarm behaviors.

\begin{figure}[thbp] 
\begin{center}
      \subfigure[]
      {\includegraphics[width=0.45\textwidth]{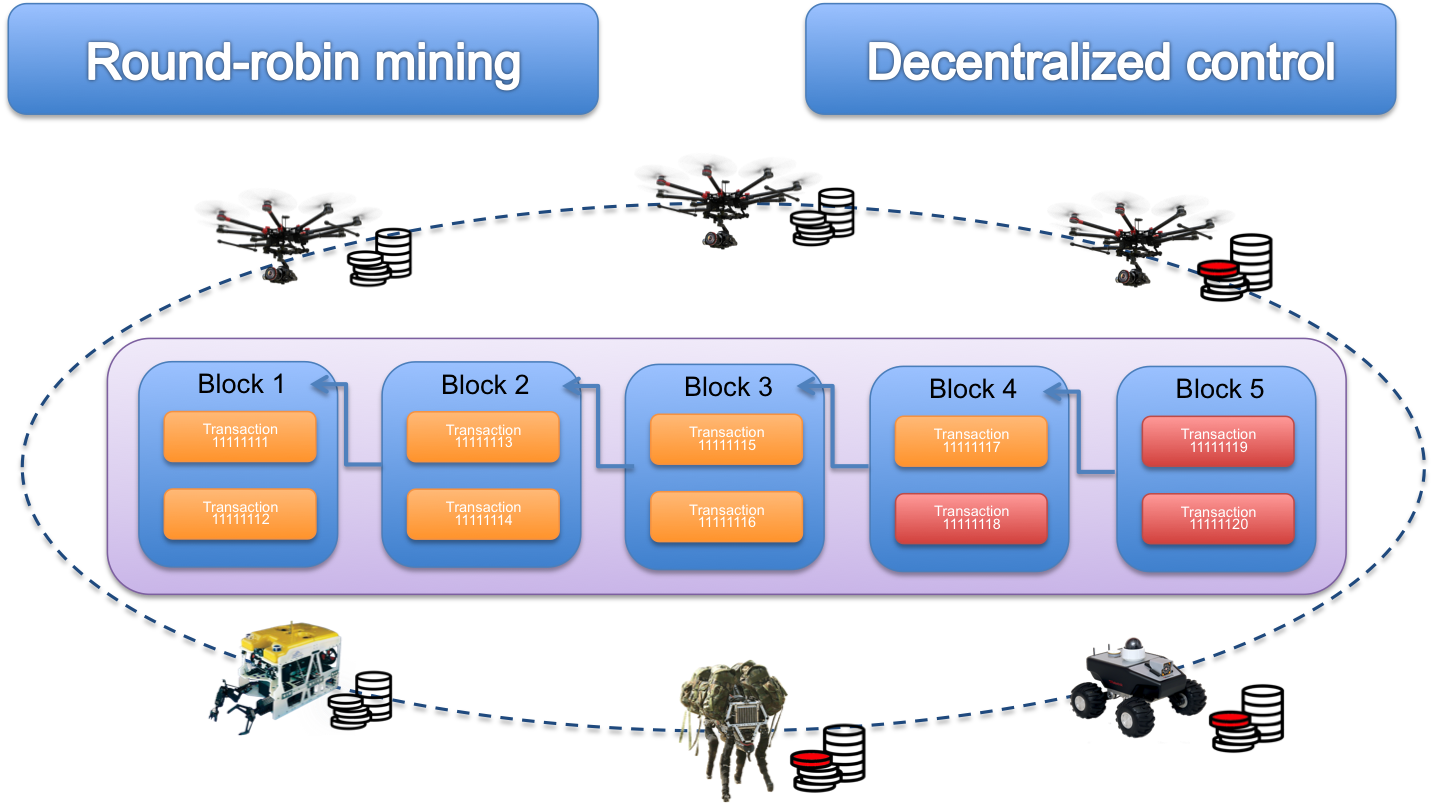}
      \label{fig:BlockChainRoundRobin}
      } 
      \subfigure[]
      {\includegraphics[width=0.45\textwidth]{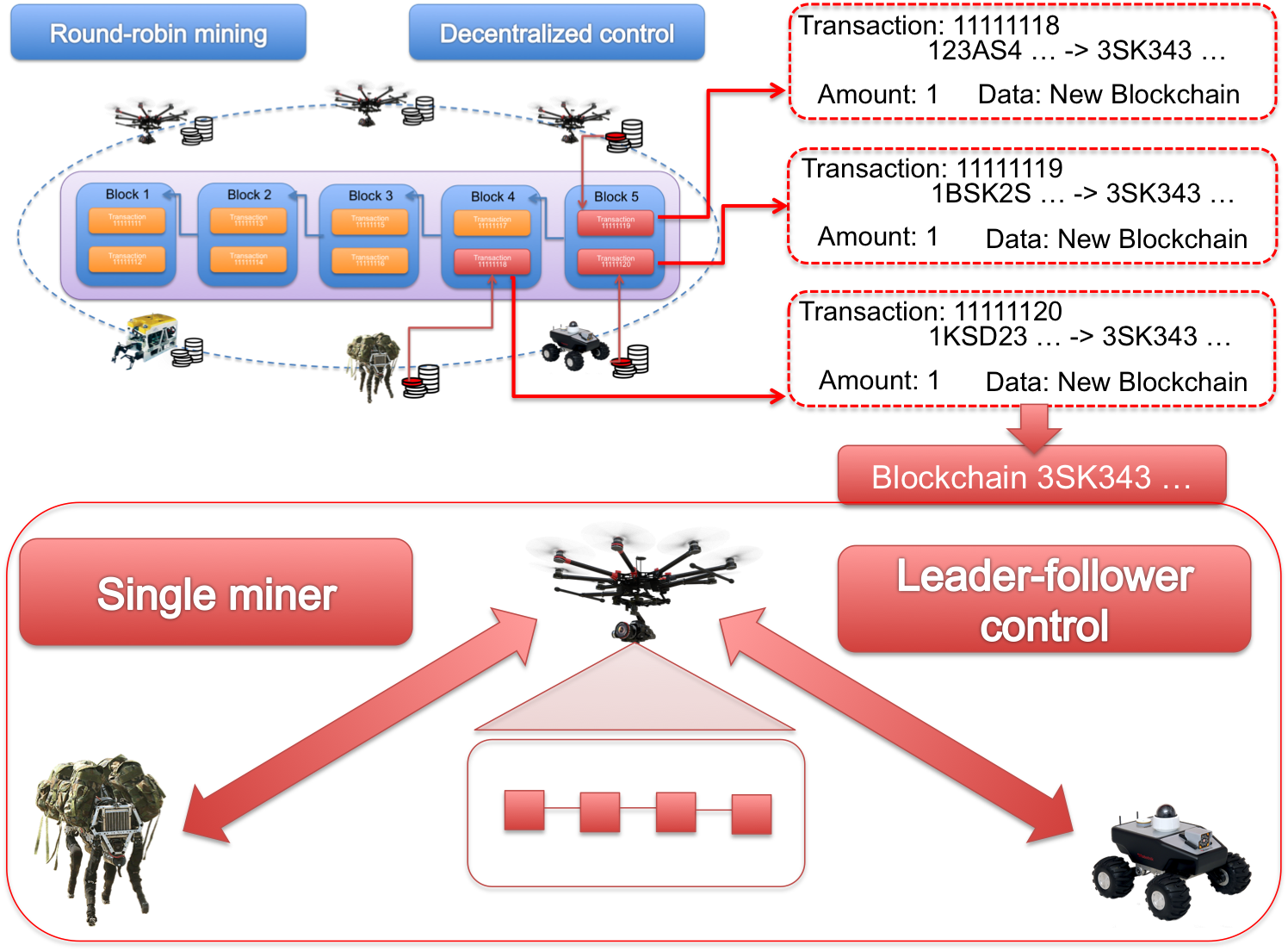}
      \label{fig:BlockChainSingleMiner}
      }
      \label{fig:BlockChainBehaviorDifferentiation} 
      \caption{(a) A typical blockchain configuration in which all
      agents in the network can become miners. This configuration
      emphasizes the decentralized control approach since all
      robots help to build the blockchain ledger. (b) Several
      agents of an already established blockchain create a
      different blockchain ledger --- sending a transaction to a
      special address --- in which the mining diversity parameter is
      changed to produce a single miner configuration. This
      configuration emphasizes a centralized approach in which only
      the miner can take control of the block creation process,
      thus transforming the blockchain into a leader-follower
      control scheme.}
\end{center}
\end{figure}

For instance, open-source projects, such as
MultiChain\footnote{\url{http://www.multichain.com/download/MultiChain-White-Paper.pdf}} in combination with pegged sidechain
algorithms\footnotemark[10]{}, provide a simple way to create multiple
blockchain ledgers connected to each other that are able to run in
parallel. Figure \(\ref{fig:BlockChainRoundRobin}\) represents a
typical blockchain configuration in which the mining diversity ---
the possibility of becoming a miner --- is distributed among
network agents using a round-robin planner. In these situations,
the control behind the decision regarding which transactions become
part of the blockchain is distributed and decentralized.

However, several members of the swarm have the possibility of
creating a parallel pegged sidechain simply by making a special
type of transaction and transferring a small portion of their
assets to the alternative chain. In this sidechain, different
parameters can be optimized to obtain a different behavior. Figure
\(\ref{fig:BlockChainSingleMiner}\) provides an overview of how a
decentralized mining scheme can be turned into a centralized mining
scheme. In the bottom part of Fig.
\(\ref{fig:BlockChainSingleMiner}\), a single agent that has monopoly
over the transactions included in the blockchain emphasizes a
leader-follower control approach instead of a completely
decentralized model. Using this approach, different robot behaviors
can be obtained using the same robot's control law, therefore, not
increasing the complexity of robot's controller.

\subsection{New business models}
\label{sec:orgheadline5}

Although this document has explored and emphasized several
blockchain applications beyond currency, it should be remembered
that blockchain technology can also be seen as an ideal Application
Programming Interface (API) for economic applications, which may
allow swarms of robots to directly take part in an economy. For
this reason, blockchain technology has the potential to stimulate
the use of swarm robotics in industrial and market-based
applications.

One of the most obvious prototypical implementations regarding the
use of robotic swarms in economic applications is the process of
exchanging data for currency between a robot and a
requester. Sensing-as-a-Service (\(S^{2}aaS\))
\cite{Perera2013,Mizouni2013,Noyen2014} is an emerging business model
pattern, which is rising in the Internet of Things (IoT)
field. \(S^{2}aaS\) helps to create multi-sided markets for sensor
data in which one or more customers --- the markets' buying side ---
subscribe to and pay for data that is provided by one or more
sensors --- selling side.

This model, which was initially designed to match the
characteristics of sensor networks distributed in smart cities and
controlled areas \cite{Zhang2015}, can be extended with the use of
robot swarms to develop more resilient and adaptive mission control
for whatever target application the user may desires.

\begin{figure}[thbp] 
\begin{center}
\includegraphics[width=0.5\textwidth]{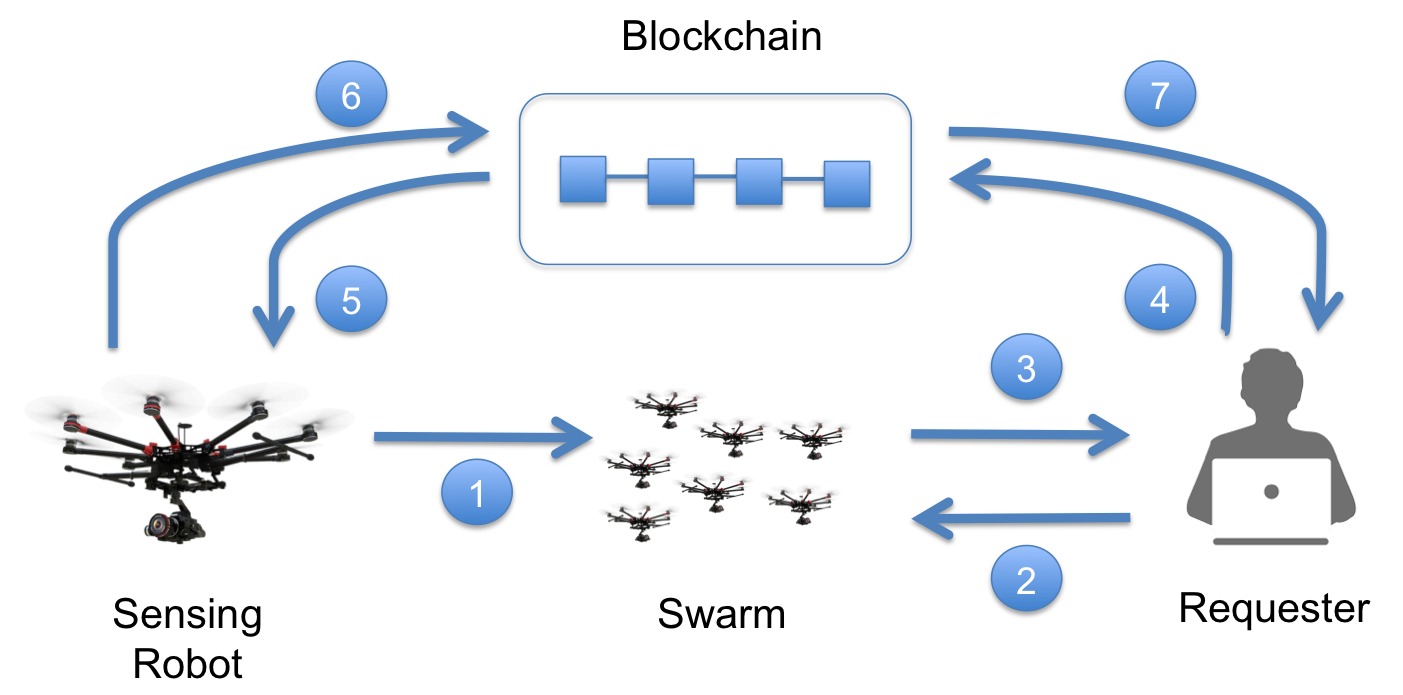}
\caption{Process of exchanging data for currency in a Sensing-as-a-Service model.}
\label{fig:BlockChainSensingAsService}                 
\end{center}
\end{figure}

Figure \(\ref{fig:BlockChainSensingAsService}\) outlines a possible
working model in which swarm robotics and blockchain technology are
combined to develop effective \(S^{2}aaS\) applications. In
Fig. \(\ref{fig:BlockChainSensingAsService}\), (1) individual robots
register into a swarm where they can be found by a
requester. Robotic swarms in this case can be regarded as a list of
addresses where additional information, including the location of
each agent, price of data provided, etc., can be found. In more
advanced scenarios, robotic swarms can even build Decentralized
Collaborative Organizations (DCOs) like those mentioned in the
first section of this document. (2) The requester can ask for a
complete list of these robots and their sensing services, (3) which
is sent back by the swarm based on the robots currently
available. If the requester decides to purchase the sensing service
provided by a specific robot, (4) he/she can send its corresponding
payment directly to the robot's public address. (5) This initial
transaction is included in the blockchain and a payment
notification is sent to the corresponding sensing robot. (6) At
this point, the hired robot can start working and send a
transaction containing the sensing data. Previous research
\cite{Zyskind2015} in privacy-oriented blockchain applications has
demonstrated that encrypting links to an off-chain site with the
requester's public key and encapsulating them in the data field of
a transaction prevents blockchain's congestion and ensures that
only the requester can read the intended message. (7) Finally, the
requester can obtain access to his/her paid data through the
transaction sent by the sensing robot.

This IoT-swarm model may be relevant for different types of private
organizations. For example, car manufacturers may require
information about road conditions, especially when bad weather
conditions arise or in disaster-relief missions where first-hand
information is crucial. Furthermore, farmers and precision
agriculture/aquaculture companies may require accurate weather
forecasts for large production areas where different types of
robots are able to provide a global view.

Finally, blockchain technology can be crucial in situations in
which multiple swarms from competitor companies have to coexist in
the same environment, such as in mining scenarios, intelligent
transportation environments, or search \& rescue missions. The
possibility that different company systems can share a secure
communication medium in which the transaction order and timestamps
are taken into account opens a path to providing a suitable
framework for competitive swarm systems.

\begin{figure}[thbp] 
\begin{center}
\includegraphics[width=0.45\textwidth]{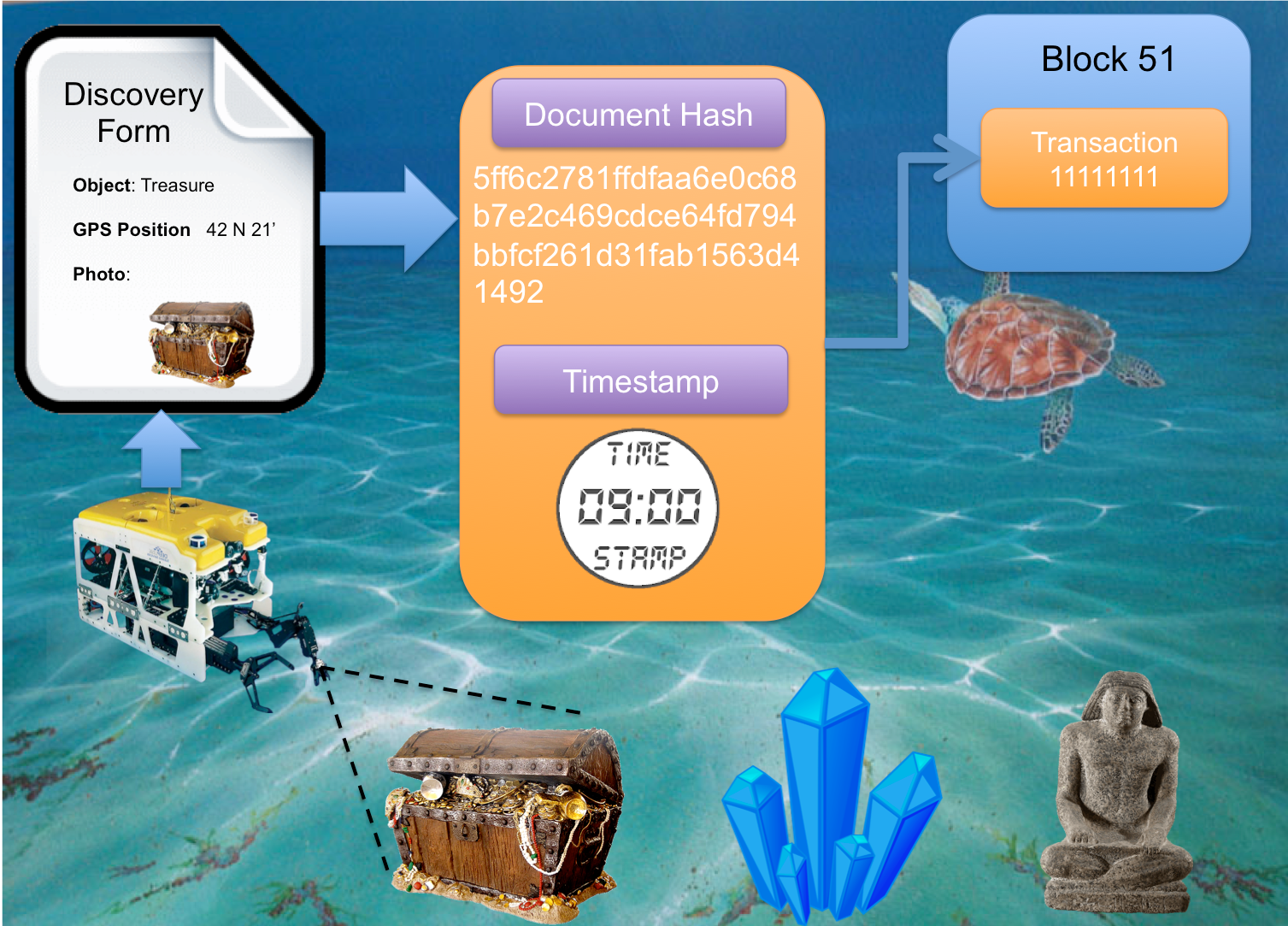}
\caption{ A UUV discovers several objects of interest --- e.g.,
treasure, mineral resources, or archaeological findings --- and files
a discovery form claiming rights over the discovered objects.}
\label{fig:BlockChainAttestation}                 
\end{center}
\end{figure}

Robotic agents that have been programmed to find resources,
objects, tokens, etc., as part of their activities may be able to
claim ownership or exploitation rights on behalf of their
owner. Figure \(\ref{fig:BlockChainAttestation}\) outlines a simple
deep seabed exploration scenario in which a UUV discovers several
objects of interest and files a discovery document claiming rights
over them. The document may contain key information about the
discovery, including the location of the discovered objects,
preliminary descriptions and reports, and even graphical
data. After calculating the document's hash, this can be included
into a blockchain's transaction and stored in the public ledger as
a proof of discovery.

This is possible due to two powerful blockchain techniques known as
hashing and time-stamping. As mentioned above, a hash string can
act as a unique and private identifier for a piece of information
or a file's contents. The hash represents the exact content of an
original piece of information, much like a digital
fingerprint. Furthermore, the hash is short enough to be included
as text in a blockchain transaction, thus providing a secure
time-stamping function of when a specific attestation transaction
occurred. Via the hashing functionality, the original document
content can be encoded into the blockchain without being
disclosed. In this way, the blockchain can be used to prove the
existence of the exact contents of a document or other digital
asset at a certain time. Whenever a proof of existence needs to be
confirmed, if the recomputed hash is the same as the original hash
registered in the blockchain, the document can be verified as
unchanged.

In this sense, blockchain technology may provide an infrastructure
for ensuring that robotic swarm systems follow specified legal and
safety regulations as they become increasingly integrated into
human society, and may result in the creation of new business
models for swarm operation.

\section{Limitations and problems to overcome}
\label{sec:orgheadline9}

Even though the combination of blockchain technology and swarm
robotics can provide useful solutions to tackle the aforementioned
issues, a number of technical challenges related to the blockchain
have been identified \cite{MelanieSwan2015} and shall require
investigation by future researchers. Solutions to these issues
might not have a direct impact on the development of new services
and businesses based on blockchain technology \emph{per se}; however,
they would be necessary steps towards mainstream adoption.

\subsection{Latency}
\label{sec:orgheadline7}

Currently, with the most widely used version of the blockchain
--- Bitcoin --- a block takes around 10 minutes to be processed. This
means that a transaction takes approximately 10 minutes to be
confirmed. Even though this rule can be modified in private
blockchains via the addition of different mining policies, such as
proof-of-stake, users in the Bitcoin network normally wait until
two or three blocks are appended to the blockchain to confirm
their transactions. This way users decrease their risk of
suffering a double-spending attack. Therefore, latency appears in
the form of the time difference between the moment a transaction
is sent and the moment it is confirmed.

The latency issue becomes highly relevant when robots are used in
formation control or cooperative tasks. In these situations, fast
and reliable information is required to orchestrate the movements
of the swarm. Collisions or other inconveniences might arise in
situations when there is a mismatch between the current state of
affairs and the one in which the transaction was originated.

Innovative research is needed to address the latency issue and to
investigate which applications are most suited for both ends of
the security vs. speed trade-off. One possible solution to
mitigate this problem might be to create affiliation-based systems
in which robots belonging to the same organization or company are
not required to wait long periods of time to accept or process
transactions among themselves. A reputation system could be
constructed from lists of previous accepted transactions within
the group to cut these waiting times. 

\subsection{Size, throughput and bandwidth}
\label{sec:orgheadline8}

If large quantities of robots are deployed for long periods of
time, they might expand the blockchain to a point where they
cannot keep a copy of the full ledger of transactions
anymore. This problem, which the Bitcoin community calls ``bloat''
\cite{Wagner2014}, is of particular importance in swarm robotics
where simple robots with limited hardware capabilities are used.

Private blockchains, such as the ones presented in this report,
are intended to have a relatively small size. However, the reality
is that if a blockchain were scaled to function in mainstream
applications, it would need to be big enough to allocate several
types of information.

Future researchers in the blockchain field have to trial different
accessibility methods to find which is the most suitable for
obtaining information from a blockchain. New interfaces such as
Chain\footnote{\url{http://chain.com/index.html}} may be able to facilitate automated calls to
a blockchain by providing address balances and balance change, as
well as notifying agents when new transactions or blocks are
created on the network.

Even though important parameters such as the block size --- how many
transactions are included in each block --- can be changed, it is
important to note that the most widely used blockchain
implementation can only handle a maximum of seven transactions per
second \cite{MelanieSwan2015}. This limitation severely compromises
the throughput of the system in busy networks with a large number
of agents. One way to tackle this issue is to raise the number of
transactions a block can contain. However, this leads to other
issues related to blockchain size and bloat. Another solution
would be to create parallel blockchains where block size and
frequency parameters are optimized for different types of
information.

\section{Conclusions}
\label{sec:orgheadline10}

Blockchain technology demonstrates that by combining peer-to-peer
networks with cryptographic algorithms, a group of agents can reach a
agreement on a particular state of affairs, and can record that
agreement in a verifiable manner without the need for a controlling
authority. Even though this technology is in its infancy, it is
already capable of extended functionalities outside its original
application, and shows promise for the creation of state-of-the-art
models in combination with other emerging technologies.

Due to the latest advances in the field, swarm robotic systems have
been gaining popularity in the last few years and are expected to
reach the market in the near future. However, several of the
characteristics that make them ideal for certain future
applications —-- robot autonomy, decentralized control, collective
emergent behavior, etc. --- hinder the evolution of the technology from
academic institutions to use in real-world problems, and eventually to
widespread industrial use.

In this work, we discussed how the combination of blockchain
technology and swarm robotic systems can provide innovative solutions
to four emergent issues, by using the robots as nodes in a network and
encapsulating their transactions in blocks. First, new security models
and methods can be implemented in order to give data confidentiality
and entity validation to robot swarms, therefore making them suitable
for trust-sensitive applications. Second, distributed decision making
and collaborative missions can be easily designed, implemented, and
carried out by using special transactions in the ledger, which enable
robotic agents to vote and reach agreements. Third, robots may be able to
function in diverse and changing environments if their operation
corresponds to different blockchain ledgers that use different
parameters, without any change in their control algorithm. In short,
these improvements would increase robots' flexibility without
increasing the complexity of the swarm design. Finally, blockchain
technology may provide an infrastructure for ensuring that robotic
swarm systems follow specified legal and safety regulations as they
become increasingly integrated into human society, and could even
result in the creation of new business models for swarm operation.

The addition of blockchain models to robotic swarms does have its
limitations, and some critics might see it as a deviation from the
minimalistic approach usually followed in swarm robotics
research. This path is subject to debate, and decisions on it must be
made in relation with the state of the art in technology. A promising
trend is recent advancements in low-power communication and processing
chips, which give advanced capabilities robots in swarm-related
activities and reduce their price, leading to the possibility of
obtaining ``more-advanced'' swarm robotic units.

In conclusion, the integration of blockchain technology could be the
key to serious progress in the field of swarm robotics. This step
could open the door not only to new technical approaches, but also to
new business models that make swarm robotics technology suitable for
innumerable market applications.

\bibliographystyle{plain}
\bibliography{References}
\end{document}